\documentclass[%
 amsmath,amssymb,
 aps,
]{revtex4}

\usepackage{graphicx, xcolor}
\usepackage{subfig}
\usepackage{float}
\usepackage{tabularx}
\usepackage[figurename=Figure]{caption}
\usepackage[tablename=Table]{caption}
\begin{document}

\preprint{APS/123-QED}

\title{Approximating Martingale Process for Variance Reduction \\
in Deep Reinforcement Learning with Large State Space}
\author{Charlie Ruan}
\affiliation{Department of Computer Science, Cornell University; \\ 
School of Operations Research and Information Engineering, Cornell University}

\date{November 2022}

\begin{abstract}
Approximating Martingale Process (AMP) is proven to be effective for variance reduction in reinforcement learning (RL) in specific cases such as Multiclass Queueing Networks. However, in the already proven cases, the state space is relatively small and all possible state transitions can be iterated through. In this paper, we consider systems in which state space is large and have uncertainties when considering state transitions, thus making AMP a generalized variance-reduction method in RL. Specifically, we will investigate the application of AMP in ride-hailing systems like Uber, where Proximal Policy Optimization (PPO) is incorporated to optimize the policy of matching drivers and customers.

\end{abstract}

\maketitle

\section{Introduction}

Ride-hailing services, such as Uber, Lyft, and Didi Chuxing, have become a popular stochastic process problem being studied in operations research. Approximating the optimal policy of matching drivers and customers in real-time is especially difficult due to the large state space and the combinatorial nature of the problem. In \cite{Feng_2021}, the authors consider a Markov decision process (MDP) model of a ride-hailing service system and innovatively decompose the MDP actions by sequentially assigning tasks to available drivers due to the large action space. Then, the reinforcement learning algorithm proximal policy optimization (PPO) \cite{schulman} is adopted for the ride-hailing system's control policy optimization.

On the other hand, Multiclass Queueing Networks (MQNs) are a special class of stochastic processing networks, a classic problem in operations research. In order to find the optimal control policy of such a network, \cite{Dai_2022} formulated the MQN problem with Poisson arrival and exponential service time as an MDP, also using reinforcement learning algorithm PPO from \cite{schulman} to optimize the network's policy. However, a naive implementation of PPO would not perform well when the network experiences heavy traffic which causes high variance. The paper \cite{Dai_2022} thus adopts several variance-reduction techniques, among which is the Approximating Martingale Process method from \cite{Hend1997}, demonstrated as an essential part of the final optimization algorithm.

In this paper, we generalize the Approximating Martingale Process (AMP) method as a variance-reduction technique in reinforcement learning, specifically under systems with a large state space and hence uncertainties in state transitions. We will consider the ride-hailing service system as the example. Intuitively, if the AMP method reduces variance effectively in the ride-hailing context, then the required number of Monte-Carlo episodes to roll-out for each policy iteration will decrease.

\section{Formulating AMP in Ride-Hailing}
In this section, we formulate the use of AMP in the ride-hailing problem under the setup of \cite{Feng_2021} by imitating the approach in Section 4.2 in \cite{Dai_2022}.

In \cite{Feng_2021}, the authors define a value function $V_\pi : S \rightarrow \mathbb{R}$ of policy $\pi$: 
\begin{equation} \label{val}
    V_\pi(s_{t,i}) := \mathbb{E}_\pi \biggl[ \sum_{k=i}^{I_t}c(s_{t,k}, a_{t,k}) + 
    \sum_{j=t+1}^{H}\sum_{k=1}^{I_j}c(s_{j,k}, a_{j,k}) \biggr],
\end{equation}
for each epoch $t = 1, ..., H$, step $i = 1, ..., I_t$, and $s_{t,i} \in S$, where $I_t$ is the number of available cars at epoch $t$, $H$ is the number of minutes in a working day (episode), and $c(s,a)$ is the reward for taking action $a$ at state $s$, following the scheme in \cite{Feng_2021}.

Then, with the roll-outs collected by Monte Carlo simulation with policy $\pi$, we can estimate the value function $V_\pi$ as
\begin{equation} \label{valOneRep}
    \hat{V}_{t,i,k} := \sum_{j=i}^{I_{t,k}}c(s_{t,j,k}, a_{t,j,k}) + 
    \sum_{l=t+1}^{H}\sum_{j=1}^{I_{l,k}}c(s_{l,j,k}, a_{l,j,k}),
\end{equation}
which is a one-replication estimate of the value function $V(s_{t,i,k})$ at state $s_{t,i,k}$, visited at epoch $t$, episode $k$, after $i-1$ steps of the ``sequential decision making process'' (SDM process).

So far, we have been revisiting the formulation in \cite{Feng_2021}. We notice that Equation (\ref{valOneRep}) may suffer from large variance as it is the sum of many random terms $c(s, a)$, the number of which depends on the step and epoch. Therefore, we focus on decreasing the magnitude of summands in Equation (\ref{valOneRep}) by following Section 4.2 in \cite{Dai_2022}. We first notice that the value function (\ref{val}) is a solution to the Bellman Equation
\begin{equation} \label{Bellman}
    V_{\pi}(s_{t,i}) = \sum_{s' \in S} P_\pi(s' | s_{t,i})V_\pi(s') +
    \mathbb{E}_{\pi(s_{t,i})} \biggl[ c(s_{t,i}, a_{t,i}) \biggr],
\end{equation}
where $P_\pi(s' | s_{t,i})$ is the probability of transitioning from state $s_{t,i}$ to state $s'$ under policy $\pi$.

Assuming that an episode $\{(s_{t,1}, a_{t,1}), (s_{t,2}, a_{t,2}), ..., (s_{t,I_t}, a_{t,I_t})\}_{t=1}^{H}$ is collected under policy $\pi$, then from the definition of Bellman Equation (\ref{Bellman}):
\[
    \mathbb{E}_{\pi(s_{t,i})} \biggl[ c(s_{t,i}, a_{t,i}) \biggr] = 
    V_{\pi}(s_{t,i}) - \sum_{s' \in S} P_\pi(s' | s_{t,i})V_\pi(s')
\]
Assuming that an approximation $\zeta$ of the value function $V_\pi$ is available and sufficiently close, then the correlation between
\[
    \mathbb{E}_{\pi(s_{t,i})} \biggl[ c(s_{t,i}, a_{t,i}) \biggr]
    \text{  and  }
    \zeta(s_{t,i}) - \sum_{s' \in S} P_\pi(s' | s_{t,i})\zeta(s')
\]
is positive, which can be exploited as the control variate to reduce the variance. Further following \cite{Dai_2022} we consider the martingale process in \cite{Hend1997} 
\begin{equation} \label{M}
    M(s_{t,i,k}) = \zeta(s_{t,i,k}) + \sum_{j=i}^{I_{t,k}} \biggl[ \sum_{s' \in S}P_\pi(s'|s_{t,j,k})\zeta(s') - \zeta(s_{t,j,k}) \biggr] + 
    \sum_{l=t+1}^{H} \sum_{j=1}^{I_{l,k}} \biggl[ \sum_{s' \in S}P_\pi(s'|s_{l,j,k})\zeta(s') - \zeta(s_{l,j,k}) \biggr].
\end{equation}
Adding $M$ to estimator (\ref{valOneRep}), we get the AMP estimator of the solution to the value function: 
\begin{align} \label{ampVal}
    \hat{V}^{AMP(\zeta)}(s_{t,i,k}) := \zeta(s_{t,i,k}) +
    & \sum_{j=i}^{I_{t,k}} \biggl[ c(s_{t,j,k}, a_{t,j,k}) + \sum_{s' \in S}P_\pi(s'|s_{t,j,k})\zeta(s') - \zeta(s_{t,j,k}) \biggr] + \notag\\
    & \sum_{l=t+1}^{H} \sum_{j=1}^{I_{l,k}} \biggl[ c(s_{l,j,k}, a_{l,j,k}) + \sum_{s' \in S}P_\pi(s'|s_{l,j,k})\zeta(s') - \zeta(s_{l,j,k}) \biggr].
\end{align}

However, we immediately notice that the $\sum_{s' \in S}P_\pi(s'|s)\zeta(s')$ is not trivial to compute for certain states $s$. We first break down the term $P_\pi(s'|s)\zeta(s')$: 
\begin{equation}
    \sum_{s' \in S} P_\pi(s'|s)\zeta(s') 
    = \sum_{s' \in S} \sum_{a \in A} \pi(a|s) P(s'|s, a)\zeta(s'),
\end{equation}
where $P(s'|s, a)$ is solely dependent on the system's dynamic. Now there are a total of three cases to consider: $(a)$ when $s$ is not the last state in an SDM, $(b)$ when $s$ is the the last state in an SDM and also the last epoch of the episode, and $(c)$ when $s$ is the last state in SDM but \textit{not} the last epoch of the episode. 

In the previous two cases, $P(s'|s, a)$ is trivial because the transition of state is deterministic when given the action. However, for the last case, when $s$ is the last state in the SDM, passenger arrivals become the randomness involved for the system's transition. Furthermore, it is computationally infeasible to iterate through all the possible passenger arrival combinations. 

This introduces to the problem that this paper will explore: whether the Approximating Martingale Process is still effective when the state space is intractably large, especially when the state transitions are impossible to iterate through.

\section{A Sampling-Based Estimated AMP in Queueing Networks}
With the problem being introduced, one intuitive approach is to approximate the AMP by sampling $L$ number of $s'$ as we cannot iterate through all the possible $s'$. In this section, we revisit the Multiclass Queueing Network \cite{Dai_2022}, and apply such a sampling-based estimated AMP. Even though the MQN case does not experience the problem of intractable number of possible transitions, we can testify whether such an AMP estimator would work by comparing its performance with the original one. Therefore, we start by modifying the original AMP estimator to a sampling-based estimated AMP.

We use Algorithm 2 in Section 4.2 of \cite{Dai_2022}, which does not involve discounting, since we are not applying discounting in the ride-hailing case. The MQN version of AMP estimator of the solution to the Poisson equation (analogous to Bellman equation in ride-hailing) is:

\begin{equation} \label{mqnAMP}
    \hat{h}_{\eta}^{AMP(\zeta)}(x^{(k)}) := \zeta(x^{(k)}) + \sum_{t=k}^{\sigma_k-1}\biggl( g(x^{(t)}) - \widehat{\mu_{\eta}^{T} g} + \sum_{y \in \chi}P_{\eta}(y|x^{(t)})\zeta(y) - \zeta(x^{(t)}) \biggr),
\end{equation}
for each state $x^{k}$, where $k = 1, ..., \sigma(N)$

Now, assuming that we do not have access to a precise $\sum_{y \in \chi}P_{\eta}(y|x^{(t)})\zeta(y)$ due to $\chi$ being impossible to iterate through, we formulate the sampling-based estimated AMP as following:
\begin{equation}\label{sampleMqnAMP}
    \sum_{y \in \chi}P_{\eta}(y|x^{(t)})\zeta(y) = 
    \sum_{y \in \chi}\sum_{a \in A} \eta(a|x^{(t)}) P(y|x^{(t)},a) \zeta(y) \approx
    \sum_{a \in A} \eta(a|x^{(t)}) \frac{1}{L} \sum_{l=1}^{L} \zeta(y_l),
\end{equation}
where $y_l$ is determined by the current state $x^{(t)}$, action $a$, and most importantly $activity_l$, which is the sampled state transition of the system. In the case of Criss-Cross Network in \cite{Dai_2022}, there are a total of 5 possible transitions: arrival of job 1 or job 2, and completion of job 1, job 2, or job 3.

\begin{figure}[H] 
    \caption{Comparison of learning curves (with $95\%$ Confidence Interval) with different sample size $L$ in Equation (\ref{sampleMqnAMP}), all using Algorithm 2 without discounting, on the Criss-Cross Network with different traffic regimes.}
    \subfloat[Imbalanced low (IL) traffic \label{samplesIL}]{%
       \includegraphics[ height=0.25\textwidth, width=0.49\textwidth]{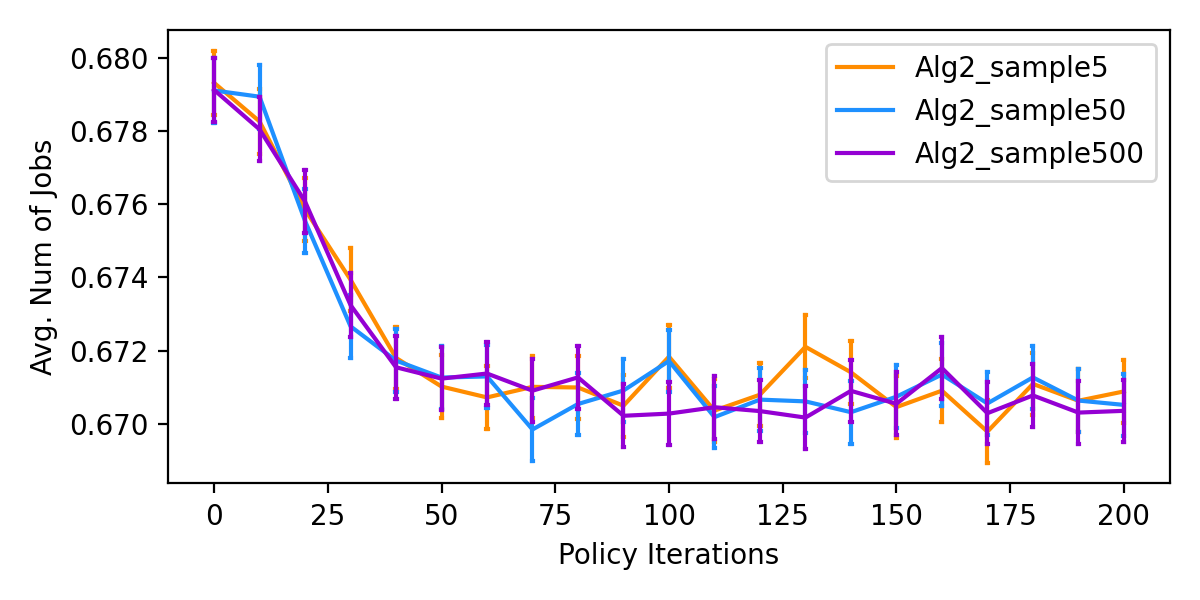}
     }
    \subfloat[Imbalanced medium (IM) traffic \label{samplesIM}]{%
       \includegraphics[ height=0.25\textwidth, width=0.49\textwidth]{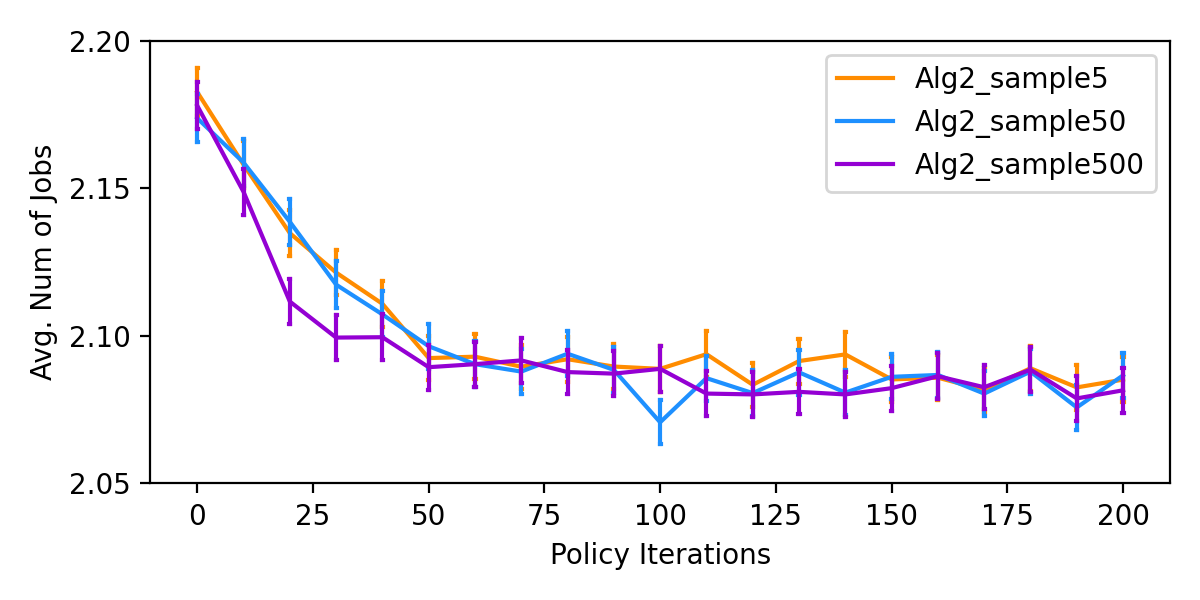}
     }\\
    \subfloat[Balanced low (BL) traffic\label{samplesBL}]{%
       \includegraphics[  height=0.25\textwidth, width=0.49\textwidth]{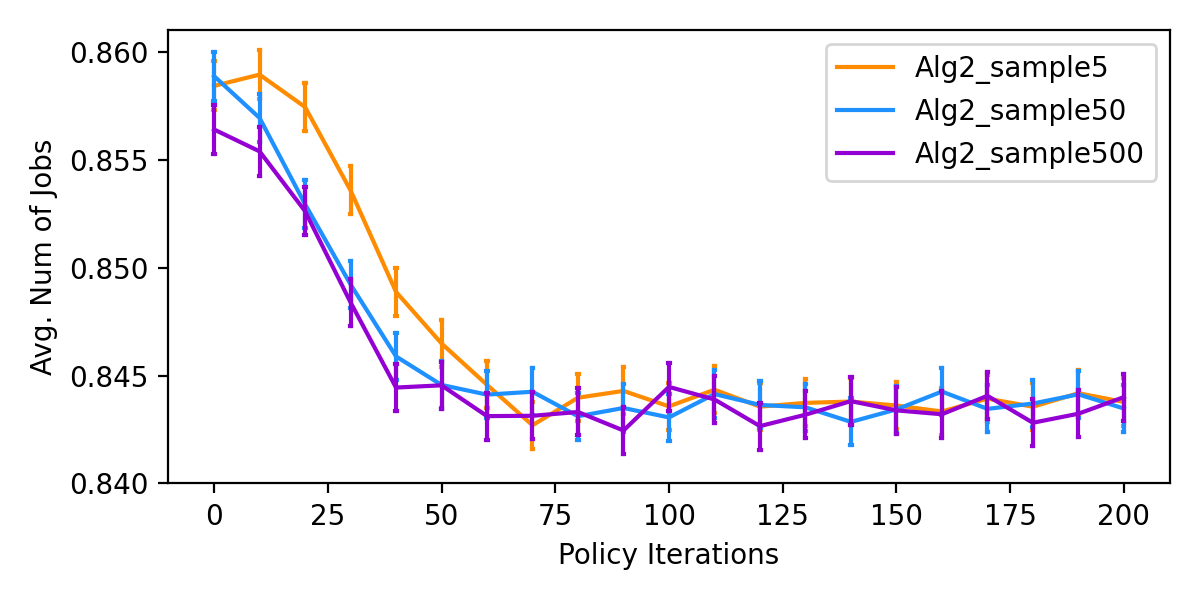}
     }
    \subfloat[Balanced medium (BM) traffic\label{samplesBM}]{%
       \includegraphics[  height=0.25\textwidth, width=0.49\textwidth]{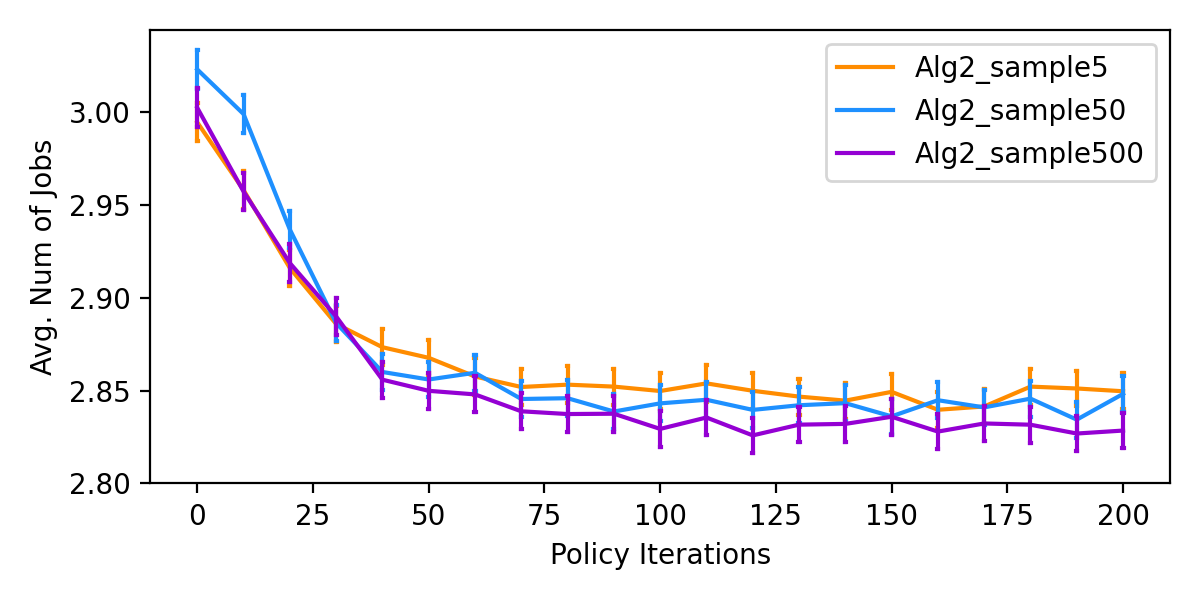}
     }
\label{fig:samplesAMPcriss}
\end{figure}

\begin{figure}[H] 
    \caption{Comparison of learning curves (with $95\%$ Confidence Interval) among Algorithm 1 (no AMP), original Algorithm 2, and Algorithm 2 using estimated AMP with a sample size of $L = 500$ on the Criss-Cross Network with different traffic regimes.}
    \subfloat[Imbalanced low (IL) traffic \label{alg12samplesIL}]{%
       \includegraphics[ height=0.25\textwidth, width=0.49\textwidth]{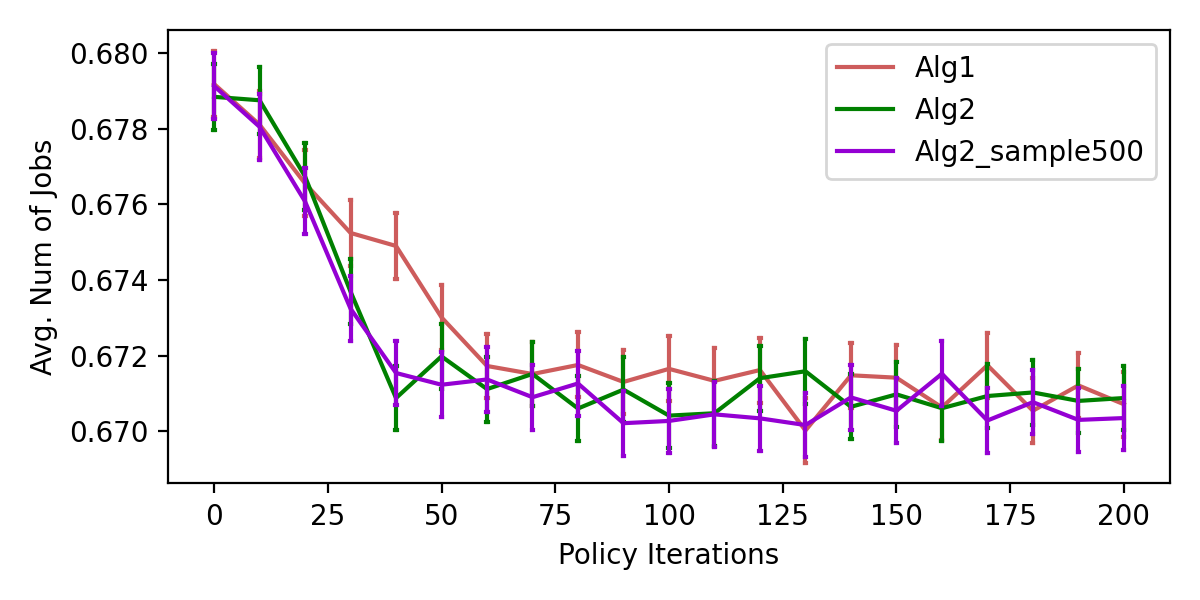}
     }
    \subfloat[Imbalanced medium (IM) traffic \label{alg12samplesIM}]{%
       \includegraphics[ height=0.25\textwidth, width=0.49\textwidth]{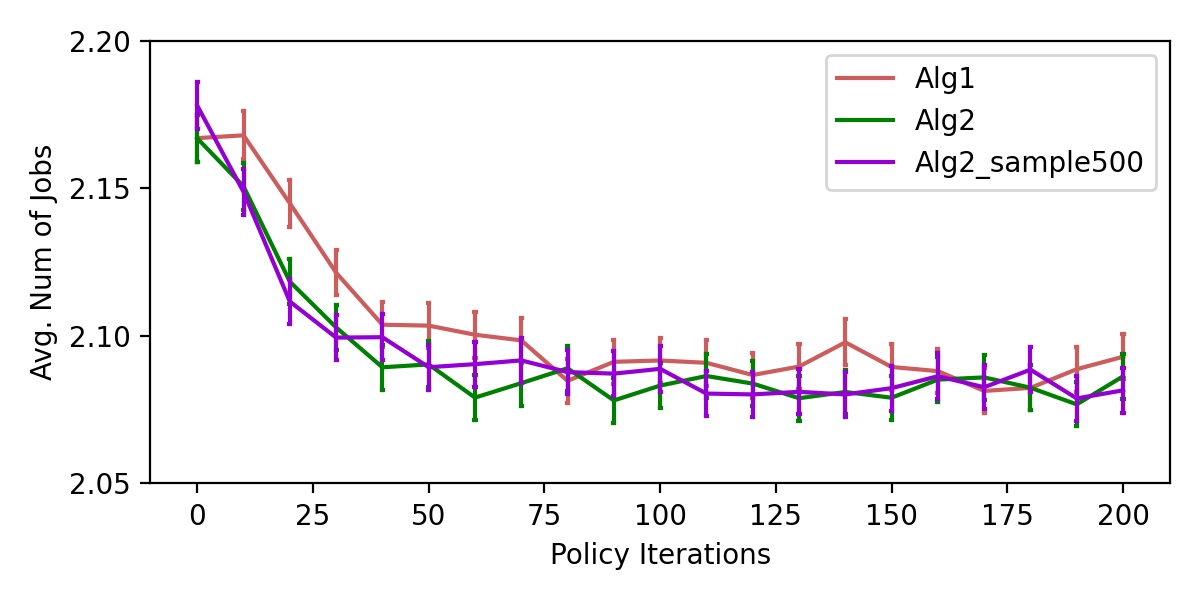}
     }\\
    \subfloat[Balanced low (BL) traffic\label{alg12samplesBL}]{%
       \includegraphics[  height=0.25\textwidth, width=0.49\textwidth]{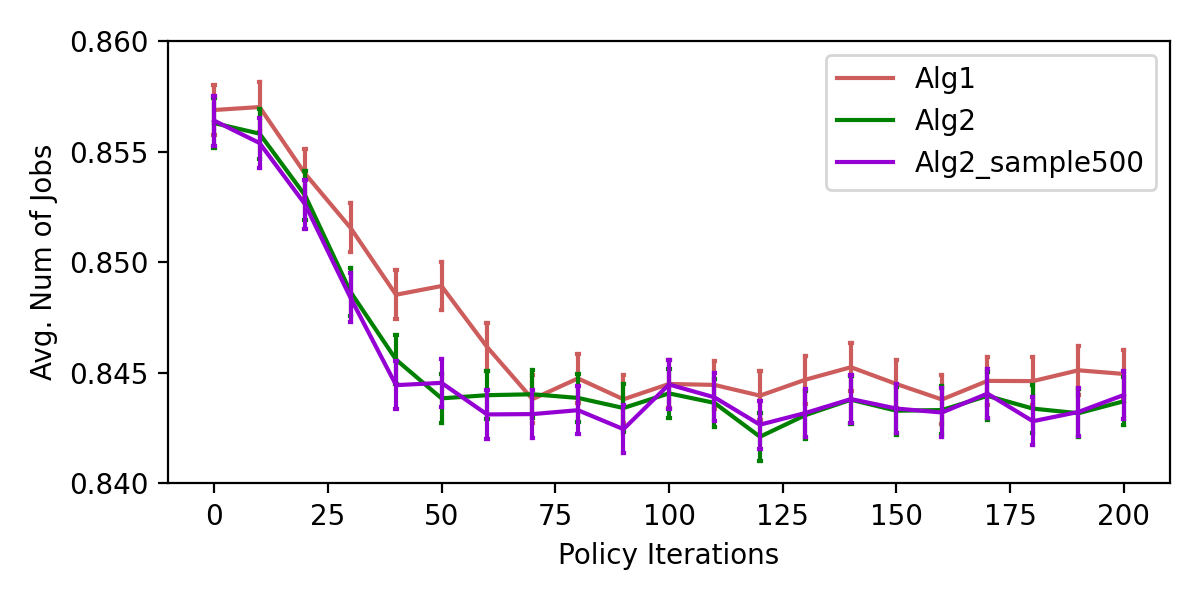}
     }
    \subfloat[Balanced medium (BM) traffic\label{alg12samplesBM}]{%
       \includegraphics[  height=0.25\textwidth, width=0.49\textwidth]{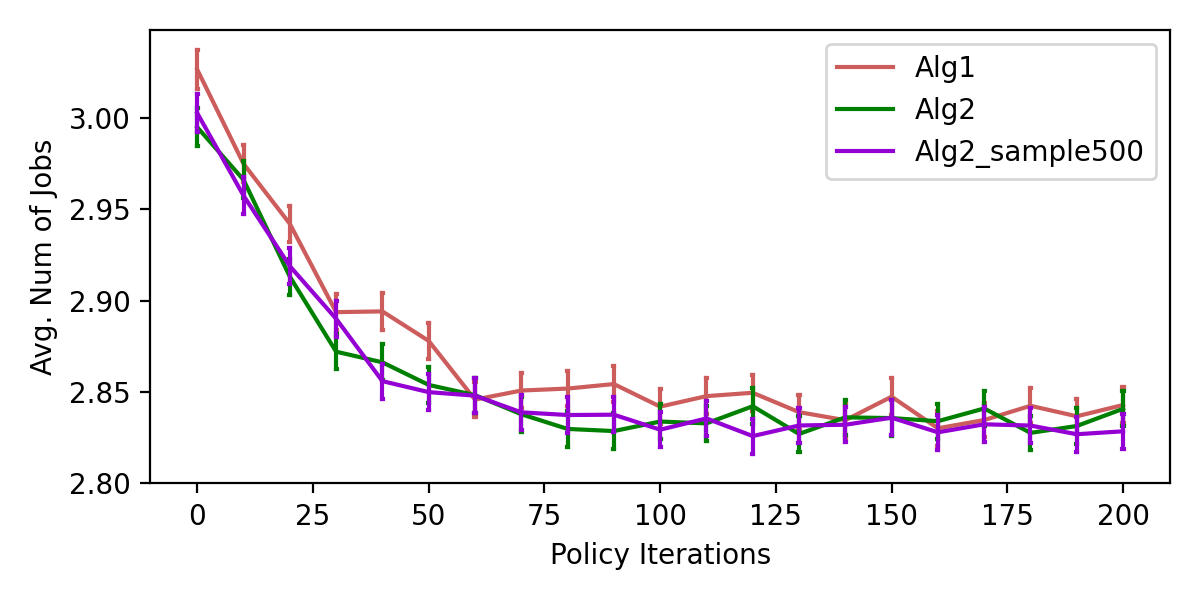}
     }
\label{fig:alg12samplesAMPcriss}
\end{figure}

As shown in Figure \ref{fig:samplesAMPcriss}, the larger the sample size $L$, the faster the convergence, which is what we would expect. Looking at Figure \ref{fig:alg12samplesAMPcriss}, we observe that the estimated AMP with a sample size of $L = 500$ has a very similar performance with the original AMP (Algorithm 2 in the original paper \cite{Dai_2022}). Therefore, we can conclude that a sampling-based estimated AMP is still effective. However, in this case, there are only 5 possible ``activities'' for the Criss-Cross network, while there are a lot more in the ride-hailing services system due to all the possible passenger arrival combinations.

\section{Current Results and Future Work}
Since the performance of Approximating Martingale Process on ride-hailing system is still inconclusive, this section documents the current progress and issues observed, along with suggestions on future efforts.

Following a similar sampling formulation, we expand the $\sum_{s' \in S} P_\pi(s'|s)\zeta(s')$ in Equation (\ref{ampVal}):

\begin{equation}\label{sampleRide}
    \sum_{s' \in S} P_\pi(s'| s_{t, i})\zeta(s') := 
    \begin{cases} 
        \sum_{a \in A}\pi(a|s_{t,i}) \frac{1}{L} \sum_{l=1} \zeta(s'_l), \text{ if } t=I_t, i \neq H, \\
        \sum_{s' \in S} \sum_{a \in A} \pi(a|s_{t,i}) P(s'|s_{t,i}, a)\zeta(s'), \text{ otherwise}
    \end{cases}
\end{equation}
where epoch $t = 1, ..., H$, step $i = 1, ..., I_t$, and sampled next state $s'_l$ is determined by current state $s_{t,i}$, action $a$, and sampled passenger arrivals. Besides, following \cite{Dai_2022}, we use the Value neural network fitted from previous policy iteration (i.e. $f_{\psi_{i-1}}$) as $\zeta$.

\begin{figure}[H]
\caption{Comparison between no AMP and estimated AMP with two different sample sizes, with Value neural network loss and car matching rates being the metrics. The experiments run with 500 cars, $H=360$ minutes, and $K=300$ episodes.}
\includegraphics[width=1\textwidth]{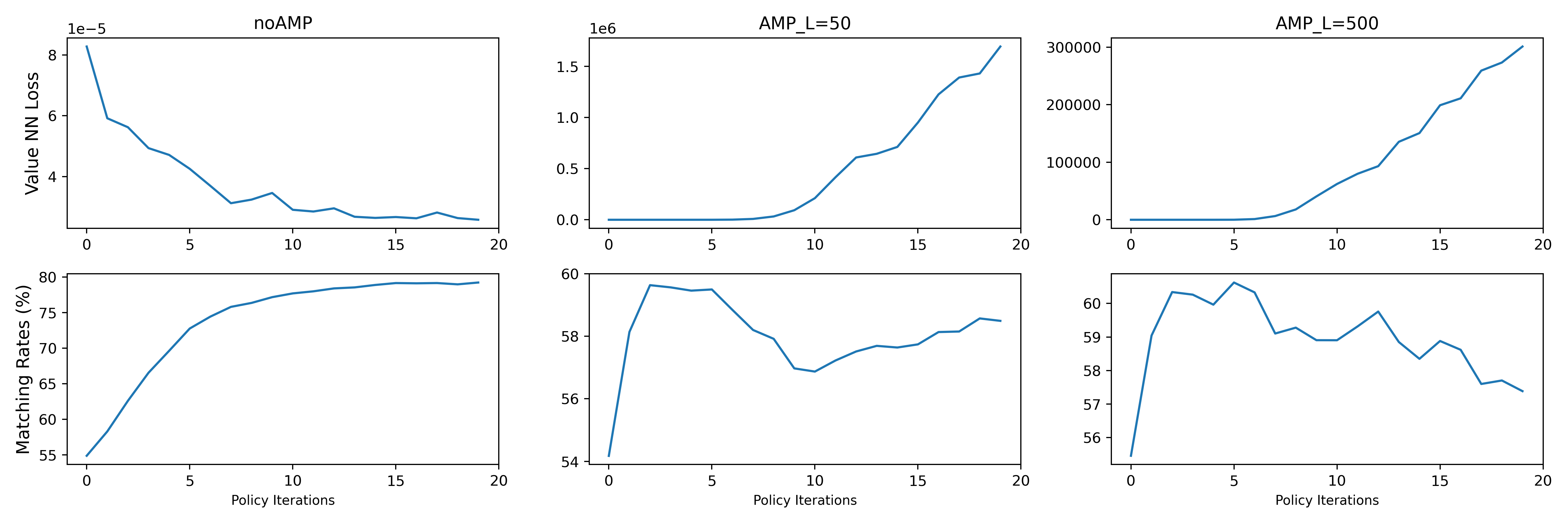}
\label{fig:rideAMPcomp}
\end{figure}

As shown in Figure \ref{fig:rideAMPcomp}, currently, such an implementation performs very badly. For both $L=50$ and $L=500$, the matching rates are not monotonously increasing as they should be, and the Value neural network's loss skyrockets after a certain epochs unlike the no-AMP counterpart. 

\begin{figure}[H]
\caption{Comparison between normalizing both input (states $s$) and output (Equation (\ref{ampVal})), and normalizing input only, with Value neural network loss and car matching rates being the metrics. The experiments run with 500 cars, $H=360$ minutes, and $K=300$ episodes.}
\includegraphics[width=1\textwidth]{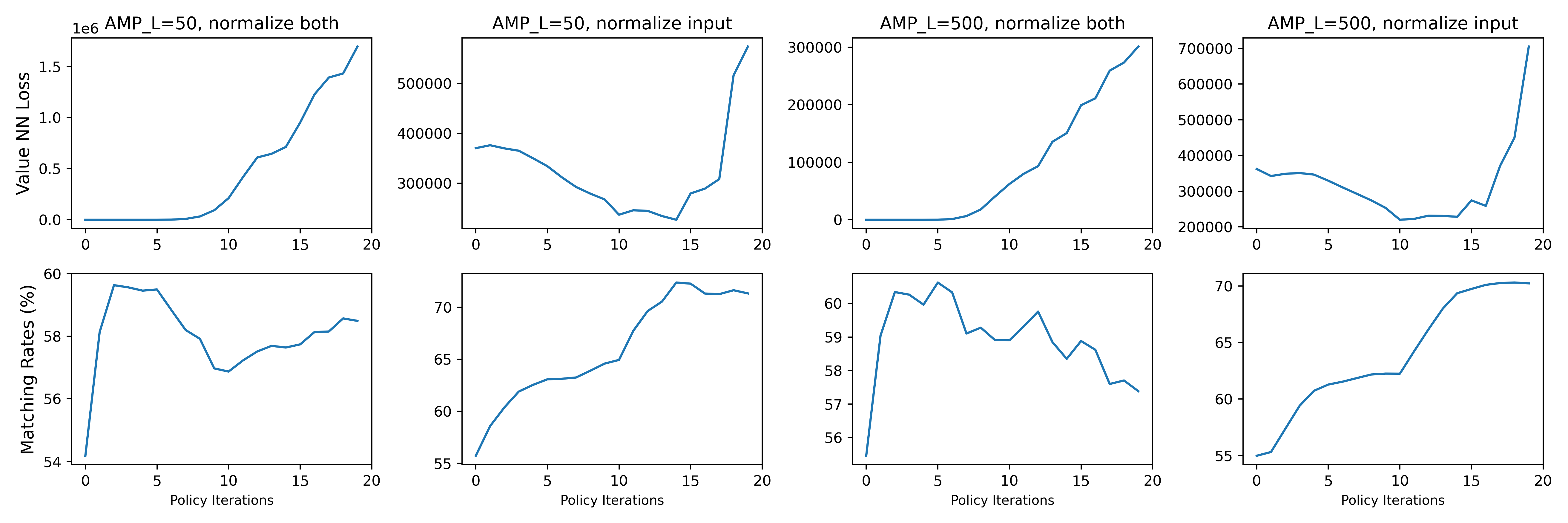}
\label{fig:rideNormComp}
\end{figure}

Based on experimental results, the way we normalize the input and output before training the Value neural network plays a huge role in the performance. In Figure \ref{fig:rideAMPcomp}, we normalize both the input (states $s$) and output (Equation (\ref{ampVal})). However, as shown in Figure \ref{fig:rideNormComp}, if we only normalize the input and not the output, we observe a better performance, though still worse than the no-AMP case. The specific rationale behind this observation is still unclear. For now, we cannot exclude the possibility of having an error hidden somewhere behind the implementation of the estimated AMP. Therefore, future work should determine whether the poor performance indeed signifies that AMP does not work on ride-hailing systems (possibly due to the large number of possible transitions as opposed to only five in Criss-Cross Network), or there was simply an error in the formulation or implementation that causes such a performance.

\section{Experimental Details}
In this section, we record relevant details about running the experiments. While the original paper \cite{Feng_2021} runs with 1000 cars and 300 episodes per policy iteration, the results in Section IV in this paper run with 500 cars and 300 episodes for efficiency, using Amazon Web Service's EC2 instance r6i.16xlarge. 

Table \ref{table:ampComp} compares the time it takes to run one iteration between no AMP and AMP with a sample size of $L=50$. In the actual implementation, we sample the next states during the Monte-Carlo simulation, explaining the time difference in the Simulation column. The difference in Data Preprocessing is due to the extra computation generated by Equation (\ref{sampleRide}). On the other hand, the sampling-based estimated AMP does not generate overhead during neural network Training. Note that Data Preprocessing accounts for computation for Equation (\ref{ampVal}) and advantage function.

Now we discuss the preferred configuration for the cloud instance to carry out the ride-hailing task. There are three general types presented by Amazon Web Service's EC2: memory-optimized instances (e.g. r6i.16xlarge), instances with high-performance GPUs (e.g. p3.8xlarge), and instances with medium-performance GPU but more CPU and RAM (e.g. g5.16xlarge). The configurations and prices for each instance are recorded in Table \ref{table:instaceComp}.

\begin{table}[H]
\centering
\caption[]{Comparison of the time it takes for one iteration between no AMP and running sampling-based AMP. Both ran on r6i.16xlarge with 500 cars and 300 episodes.}\label{table:ampComp}
\begin{tabular}{|c|c|c|c|c|}
    \hline
    Task  & Simulation (mins) & Data Preprocessing (mins) & NN Training (mins) & Total Time (mins) \\\hline
    no AMP & 3.39 & 2.58 & 14.97 & 20.94 \\\hline
    AMP with $L=50$ & 9.26 & 3.69 & 14.21 & 27.16 \\
    \hline
\end{tabular}
\end{table}

\begin{table}[H]
\centering
\caption[]{Comparison of AWS EC2 instances that we consider for the ride-hailing task. While p3 instances uses NVIDIA Tesla V100 GPUs and Broadwell E5-2686 v4 for CPUs, g5 instances use NVIDIA A10G Tensor Core GPUs and 2nd generation AMD EPYC processors for CPUs, and r6i uses Ice Lake 8375C for CPUs. Prices are recorded as of November 2022.}\label{table:instaceComp}
\begin{tabular}{|c|c|c|c|c|c|}
    \hline
    Instance  & vCPUs & Mem (GiB) & GPUs & GPU Memory (GiB) & Price (USD/hr) \\\hline
    r6i.16xlarge & 64 & 512 & N/A & N/A & 4.032 \\\hline
    r6i.24xlarge & 96 & 768 & N/A & N/A & 6.048 \\\hline
    g5.16xlarge  & 64 & 256 & 1 & 24 & 4.096 \\\hline
    g5.24xlarge  & 96 & 384 & 4 & 96  & 8.144 \\\hline
    p3.8xlarge   & 32 & 244 & 4 & 64  & 12.24 \\\hline
    p3.16xlarge  & 64 & 488 & 8 & 128 & 24.48 \\
    \hline
\end{tabular}
\end{table}

\begin{table}[H]
\centering
\caption[]{Comparison of the time it takes for one iteration between r6i.16xlarge and g5.16xlarge, both ran without AMP. We used 300 cars and 100 episodes since g5.16xlarge does not have enough RAM to finish a 500-car and 300-episode task.}\label{table:gpuComp}
\begin{tabular}{|c|c|c|c|c|}
    \hline
    EC2 Instance  & Simulation (mins) & Data Preprocessing (mins) & NN Training (mins) & Total Time (mins) \\\hline
    r6i.16xlarge & 0.63 & 0.40 & 2.32 & 3.35 \\\hline
    g5.16xlarge & 5.38 & 0.32 & 6.78 & 12.48 \\
    \hline
\end{tabular}
\end{table}

Table \ref{table:gpuComp} compares the time it takes to run one iteration of no-AMP task between using r6i.16xlarge (CPUs only) and g5.16xlarge (has 1 GPU, but less memory). Surprisingly, with a GPU, we not only have slower simulation (since simulation is a sequential and hence CPU-intensive workload), we also have a significantly slower neural network training speed. This is probably due to how g5.16xlarge's GPU only has 24 GiB Memory, since tensor computation takes place on the GPU in order to experience the speed-up that a GPU brings, and a small amount of GPU memory may slow things down. When running with 500 cars with 100 episodes (not even 300 episodes), we even run into out-of-memory (OOM) problem for the g5.16xlarge's GPU during neural network fitting. 

Then, one may think that we should use instances like p3.8xlarge to benefit from the higher GPU memory. However, even though GPUs may increase the efficiency for neural network training, the number of vCPUs and the RAM should be the priority because of the nature of our task. We need to simulate 300 episodes, thus the more vCPUs, the more parallel we can be. Besides, 300 episodes of data is not trivial: we run into out-of-memory (OOM) issues (CPU RAM in this case) on g5.16xlarge during simulation, hence Table \ref{table:gpuComp} runs with 300 cars and 100 episodes.

Upgrading the instance to g5.24xlarge or p3.16xlarge is another possible option, but AWS requires users to have sufficient historical usage with other lower-tier instances in order to apply for such instances. Besides, we cannot guarantee if the 384 GiB memory for g5.24xlarge would be enough to carry out a 500-car and 300-episode task (since r6i.16xlarge has 512 GiB). Note that if we want to run experiments with 1000 cars, instead of 500 cars, with 300 episodes, we need at least r6i.24xlarge, otherwise r6i.16xlarge might also run into out-of-memory problems.

The ideal cloud instance for the ride-hailing task is one that has the vCPUs that match the calibre of the r6i.16xlarge, along with GPUs with enough memory on top of that (maybe the four from the p3 instances). Unfortunately, there is not such an instance to my knowledge; therefore we use CPU-only instances to carry out the experiments. 

\pagebreak

\begin{acknowledgments}
First and foremost, I would like to thank Professor Jim Dai for this precious research opportunity and his exceptional guidance. Furthermore, the current progress would not have been possible without the help of Jiekun Feng and Mark Gluzman along the way. Finally, the generous funding from the School of Operations Research and Information Engineering at Cornell University was crucial for securing the computational resources.
\end{acknowledgments}

\bibliography{main}

\end{document}